\documentclass{bmvc2k}

\usepackage[inline]{enumitem}
\usepackage{graphicx}
\usepackage{caption}
\usepackage{subcaption}
\usepackage{tablefootnote}

\usepackage{tikz}
\usetikzlibrary{calc,shapes,arrows,decorations.text,positioning,patterns,matrix,chains,decorations.pathreplacing}

\usepackage{threeparttable}

\usepackage[dvipsnames]{xcolor}

\title{Improving Image Clustering With Multiple Pretrained CNN Feature Extractors}

\addauthor{Joris Gu\'erin}{jorisguerin.research@gmail.com}{1}
\addauthor{Byron Boots}{https://www.cc.gatech.edu/~bboots3}{2}

\addinstitution{
Laboratoire d'Ing\'enierie des Syst\`emes Physiques Et Num\'eriques \\
Arts et M\'etiers ParisTech \\
Lille, France
}
\addinstitution{
School of Interactive Computing\\
Georgia Institute of Technology \\ 
Atlanta, USA
}

\runninghead{Gu\'erin, Boots}{Image Clustering with Multiple Pretrained CNNs}

\begin{document}

\maketitle

\begin{abstract}
For many image clustering problems, replacing raw image data with \emph{features} extracted by a pretrained convolutional neural network (CNN), leads to better clustering performance. However, the specific features extracted, and, by extension, the selected CNN architecture, can have a major impact on the clustering results. In practice, this crucial design choice is often decided arbitrarily due to the impossibility of using cross-validation with unsupervised learning problems. 
However, information contained in the different pretrained CNN architectures may be complementary, even when pretrained on the same data. To improve clustering performance, we rephrase the image clustering problem as a multi-view clustering (MVC) problem that considers multiple different pretrained feature extractors as different ``views'' of the same data. 
We then propose a multi-input neural network architecture that is trained end-to-end to solve the MVC problem effectively. 
Our experimental results, conducted on three different natural image datasets, show that: \begin{enumerate*} \item using multiple pretrained CNNs jointly as feature extractors improves image clustering; \item using an end-to-end approach improves MVC; and \item combining both produces state-of-the-art results for the problem of image clustering. \end{enumerate*}

\end{abstract}

\section{Introduction}

Image Clustering (IC) is a major research topic in machine learning which attempts to partition unlabeled images based on their content. It has been used to solve numerous problems including web-scale image clustering~\cite{webscale}, story-line reconstruction from streams of images~\cite{storyline}, and medical image annotation~\cite{alternating_opt_clust}. In this paper, we focus on the IC setting where the number of clusters is known in advance. 

The first successful methods for IC focused on feature selection and used sophisticated algorithms to handle complex features~\cite{gmm,commonality_clustering}. 
Recently, research in IC has shifted towards using features extracted from Convolutional Neural Networks (CNN) pretrained on ImageNet~\cite{imagenet}. In \cite{infinite_ensemble,alternating_opt_clust,webscale,imsat}, pretrained CNN architectures are used to extract features from images before clustering.
As shown in \cite{supervised_transfer}, there exist a variety of publicly available pretrained CNNs that are able to generate linearly separable latent spaces for many datasets. 
For unsupervised tasks, the choice of a good pretrained architecture cannot be cross-validated and thus is often arbitrary (\cite{infinite_ensemble,alternating_opt_clust,webscale,imsat}). This is potentially problematic as~\cite{transfer_cnn_clustering} shows that the choice of architecture has a major impact on the clustering results. 

In this paper, we aim to remove the need for this design choice. Following the intuition that different pretrained deep networks may contain complementary information (see Section~\ref{sec:intuition}), we propose to use multiple pretrained networks to generate multiple feature representations. Such representations are treated as different ``views'' of the data, thus casting the initial IC problem into Multi-View Clustering (MVC). The success of ensemble methods for clustering~\cite{ensemble_survey} suggests that such an approach can improve overall clustering results. Finally, building on the recent success of end-to-end clustering methods~\cite{survey_e2e}, we also propose a parallel feed-forward neural network architecture which allows us to solve the MVC problem within existing deep clustering frameworks.

\vspace{-2pt}
\subsection{Contributions}\label{sec:contrib}

We propose to transform IC into MVC by extracting features from several different pretrained CNNs. This removes the crucial design choice of feature extractor selection. We also propose a deep learning approach to address MVC.  Our experimental results suggest:
\vspace{-1pt}
\begin{itemize}
\setlength\itemsep{0.01em}
    \item \emph{Image clustering can be improved by using features extracted from several pretrained CNN architectures, eliminating the need to select just one.}
    \item \emph{Multi-view clustering can be improved by adopting end-to-end training.}
    \item \emph{These two ideas can be combined to obtain state-of-the-art results at image clustering.}
\end{itemize}
\vspace{-1pt}
We emphasize that the two steps of the proposed methods can be used independently. Generating multiple feature representations using several pretrained CNNs can be combined with any multi-view clustering algorithm. Similarly, the proposed architecture can be leveraged to solve any MVC problems end-to-end.

\vspace{-1pt}
\section{From Image Clustering to Multi-View Clustering}

\subsection{Related work}

Ensemble clustering (EC) combines different clustering results to obtain a final partition of the original data with improved  quality~\cite{ensemble_survey}. EC is composed of two steps: generation, which creates a set of partitions, and consensus, which integrates partitions into a better set of clusters. In contrast to EC, Multi-View Clustering (MVC), is concerned with finding a unified partition from multi-view data~\cite{survey_mvc}, which can be obtained by various sensors or represented with different descriptors. Recently, MVC has received a lot of attention. In \cite{multiview_concat}, the authors propose different loss functions applied on the concatenated views, in \cite{multiview_latent1} and \cite{multiview_latent2} lower-dimensional subspaces are learned before clustering with standard methods.

MVC and EC are closely related and have been combined in prior work. In \cite{multiview_ensemble}, good MVC results are attained by embedding MVC within the EC framework. The authors leveraged the different views to generate independent partitions and then used a co-association-based method to obtain consensus. In both \cite{generated_mv_images} and \cite{generated_mv_genes}, generation mechanisms borrowed from EC are used to generate multiple artificial views of the data. In this paper, we propose to use multiple pretrained CNNs to generate different feature representations of an image dataset. Hence, we generate a MVC problem from an ensemble of pretrained CNN feature extractors.

\subsection{Intuition}\label{sec:intuition}

In \cite{transfer_cnn_clustering}, the authors show that different CNN feature extractors, pretrained on the same task (ImageNet classification), perform differently on a new target IC task. The best feature extractor does not always correspond to the best performing CNN on ImageNet, and there is no network which is consistently the best across different IC tasks. The discrepancy between the results of different methods usually motivates the use of ensemble methods.

In this case, as the different CNNs are pretrained on the same source task, the fact that they might contain complementary information for a target IC task might seem counter-intuitive. Indeed, one can expect that all networks have learned the same information. However, the task of ImageNet classification is so complex that it seems likely that there exist different latent spaces that can be leveraged to solve it. Consider the following contrived example: to recognize a car, one network might learn a wheel detector while another one might detect wing mirrors. Both sets of discriminative features would enable the solution to the ImageNet classification task, but would also carry very different information that might be useful in solving a new IC task.

\begin{figure}[!ht]
    \centering
    
    \begin{subfigure}[b]{0.3\textwidth}
        \centering \includegraphics[width=\textwidth]{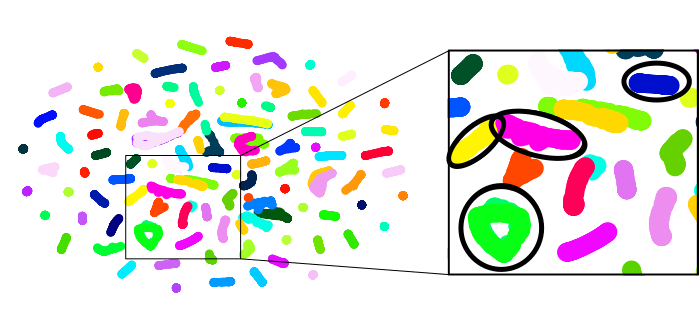}
        \caption{Xception}\label{fig:intuition_xce}
    \end{subfigure}
    ~
    \begin{subfigure}[b]{0.33\textwidth}
        \centering \includegraphics[width=\textwidth]{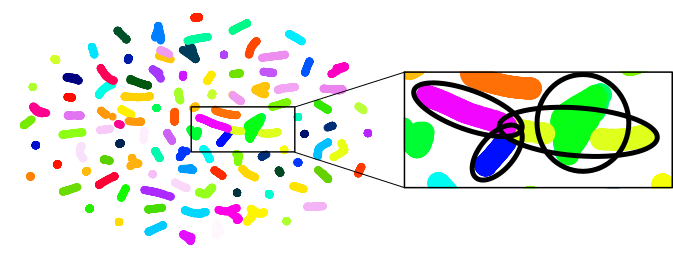}
        \caption{ResNet50}\label{fig:intuition_resnet}
    \end{subfigure}
    ~
    \begin{subfigure}[b]{0.3\textwidth}
        \centering \includegraphics[width=\textwidth]{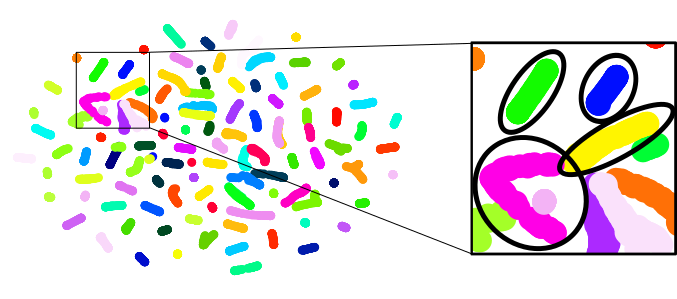}
        \caption{VGG19}\label{fig:intuition_vgg}
    \end{subfigure}
    
    \caption{\textit{Best viewed in color}. 2D t-SNE  visualization of features extracted by the last layer of three pretrained CNNs for the COIL100 dataset. These features form different \emph{complementary} views of the data.}
    \label{fig:intuition}
\end{figure}

This intuition is visualized on real data in Figure~\ref{fig:intuition}, which shows the 2D t-SNE representations~\cite{tsne} of the COIL100 dataset~\cite{coil100} for different feature extractors. Although our experimental results show that features extracted with ResNet50 present better clustering results on COIL100 (see Section~\ref{sec:exp}), we can see that both VGG19 and Xception can better separate the circled classes. Hence, it may be expected that using these three feature representations as different views of the COIL100 dataset would help improve the clustering results on the final partition.

\vspace{-5pt}
\subsection{IC problem reformulation}

Let $\mathcal{I} = \{I_1, ... I_{n}\}$ be a set of $n$ unlabeled natural images, and let $\mathcal{FE} = \{FE_1, ... FE_m\}$ be a set of $m$ feature extractors. In theory, $\mathcal{FE}$ can be composed of any function mapping raw pixel representations to lower-dimensional vectors, but in practice, we use pretrained deep CNNs. The first step in our approach is to generate a set of feature vectors from each element of $\mathcal{FE}$. $\forall i \in [1, ... m]$, we denote the matrix of features representing $\mathcal{I}$ as $V_i$,  such that it's row $V_{i,k}$ is the feature vector representing $I_k$ and extracted by $FE_i$:
\begin{equation}
\label{eq:feat}
    V_{i,k} = FE_i(I_k).
\end{equation}
$\mathcal{V} = \{V_1, ... V_m\}$ can be interpreted as a set of views representing the dataset. Thus, $\mathcal{V}$ is a multi-view dataset representing $\mathcal{I}$. The problem of clustering $\mathcal{V}$ is a MVC problem, which can be solved using any MVC algorithm~\cite{survey_mvc}. A visual representation of the multiview generation mechanism can be seen in the blue frame of Figure~\ref{fig:method}.

\newlength{\layerspace} \setlength{\layerspace}{0.2\textwidth}

\tikzset{sum/.style = {draw, circle, node distance = 2cm}}
\newcommand{\suma}{\Huge$+$}

\begin{figure}[!ht]
    \centering
    \scalebox{0.9}{
        \begin{tikzpicture}
            \node[inner sep=0pt] (im) at (0,0)
            {\includegraphics[width=.15\textwidth]{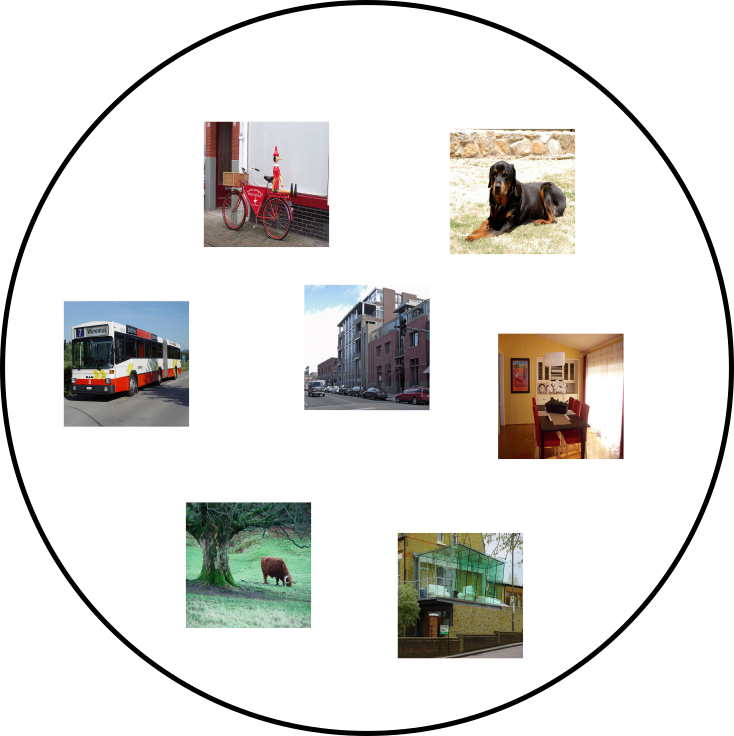}};
            \node[inner sep=0pt, font = \bf] (im_txt1) at (0, 1.8) {Unsupervised};
            \node[inner sep=0pt, font = \bf] (im_txt2) at (0, 1.45) {image set};
            
            \node[inner sep=0pt] (vgg) at (\layerspace, 1.5)
            {\includegraphics[width=.15\textwidth, height=.06\textwidth]{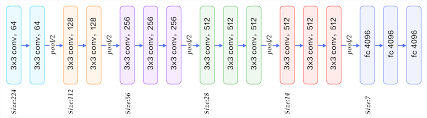}};
            \draw (vgg.north east) rectangle (vgg.south west);
            
            \node[inner sep=0pt] (res) at (\layerspace, 0.5)            {\includegraphics[width=.15\textwidth, height=.06\textwidth]{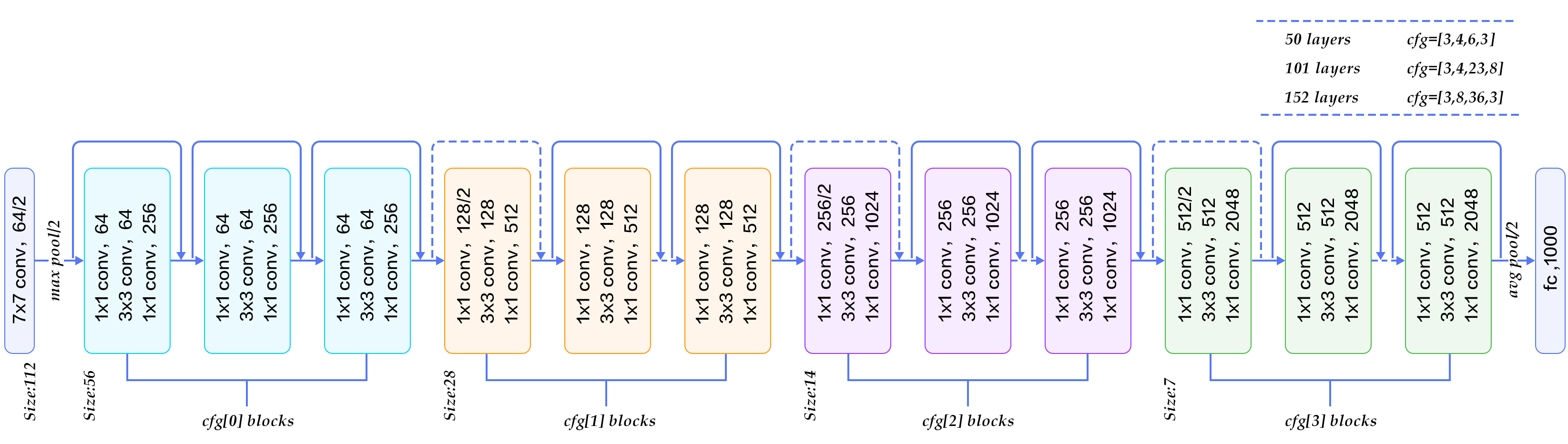}};
            \draw (res.north east) rectangle (res.south west);
            
            \node[inner sep=0pt] (inc) at (\layerspace, -1.5)
            {\includegraphics[width=.15\textwidth, height=.06\textwidth]{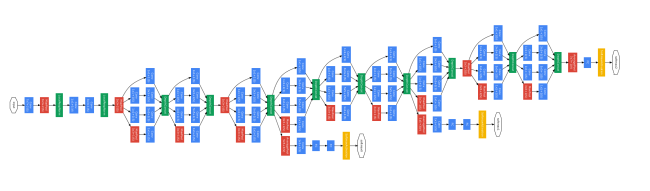}};
            \draw (inc.north east) rectangle (inc.south west);
            
            \node[inner sep=0pt, font = \bf] (feat_txt1) at (\layerspace, 3) {Deep feature};
            \node[inner sep=0pt, font = \bf] (feat_txt2) at (\layerspace, 2.65) {extractors};
            
            \draw[darkgray,thick,dotted] ($(vgg.north east)+(0.15,0.15)$) rectangle ($(inc.south west)+(-0.15,-0.15)$);
            \node[darkgray, above = 0.1 of vgg.north] (a) {$FE$};
            
            \path (res) -- (inc) node [font=\Huge, midway, sloped] {$\dots$};
            
            \draw[blue,thick,dotted] ($(inc.south east)+(0.5,-0.4)$) rectangle ($(im.north west)+(-0.4,2.3)$);
            \node[blue, above = 2.3cm of im.north east] (a) {Multi-view generator};
            
            \node[inner sep=0pt] (fc1) at (2.1\layerspace, 1.5)
            {\includegraphics[width=.07\textwidth, height=.06\textwidth]{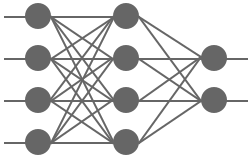}};
            \draw (fc1.north east) rectangle (fc1.south west);
            
            \node[inner sep=0pt] (fc2) at (2.1\layerspace, 0.5)
            {\includegraphics[width=.07\textwidth, height=.06\textwidth]{fc.png}};
            \draw (fc2.north east) rectangle (fc2.south west);
            
            \node[inner sep=0pt] (fc3) at (2.1\layerspace, -1.5)
            {\includegraphics[width=.07\textwidth, height=.06\textwidth]{fc.png}};
            \draw (fc3.north east) rectangle (fc3.south west);
            
            \node[inner sep=0pt, font = \bf] (clus_txt1) at (2.1\layerspace, 3) {Multi-layer};
            \node[inner sep=0pt, font = \bf] (clus_txt2) at (2.1\layerspace, 2.65) {perceptrons};

            \draw[darkgray,thick,dotted] ($(fc1.north east)+(0.15,0.15)$) rectangle ($(fc3.south west)+(-0.15,-0.15)$);
            \node[darkgray, above = 0.1 of fc1.north] (a) {$mlp$};
            
            \path (fc2) -- (fc3) node [font=\Huge, midway, sloped] {$\dots$};
            
            \node at  (2.8\layerspace, 0) [sum, name=concat] {\suma};
            
            \node[inner sep=0pt] (fcf) at (3.5\layerspace, 0)
            {\includegraphics[width=.07\textwidth, height=.06\textwidth]{fc.png}};
            \draw (fcf.north east) rectangle (fcf.south west);
            
            \node[darkgray, inner sep=0pt, font = \bf] (fcf_txt) at (3.5\layerspace, 0.8) {$mlp_{out}$};
                        
            \draw[red,thick,dotted] ($(fcf.north east)+(0.3,2.9)$) rectangle ($(fc3.south west)+(-0.5,-0.4)$);
            \node[red, above = 3.1cm of concat.north] (a) {Deep multi-view clustering network};
            \node[red, above = 2.7cm of concat.north] (a) {(MVnet)};
            
            \node[inner sep=0pt, font = \bf] (lab) at (2.8\layerspace, 1) {\small Concatenate};
            
            \node[inner sep=0pt, font = \bf] (lab) at (4.2\layerspace, 0) {Clusters};
            
            \draw[->, line width = 0.3mm] (im.east) -- (vgg.west);
            \draw[->, line width = 0.3mm] (im.east) -- (res.west);
            \draw[->, line width = 0.3mm] (im.east) -- (inc.west);
            
            \draw[->, line width = 0.3mm] (vgg.east) -- (fc1.west);
            \draw[->, line width = 0.3mm] (res.east) -- (fc2.west);
            \draw[->, line width = 0.3mm] (inc.east) -- (fc3.west);
            
            \draw[->, line width = 0.3mm] (fc1.east) -- (concat.west);  
            \draw[->, line width = 0.3mm] (fc2.east) -- (concat.west);  
            \draw[->, line width = 0.3mm] (fc3.east) -- (concat.west);  
            
            \draw[->, line width = 0.3mm] (concat.east) -- (fcf.west);  
            
            \draw[->, line width = 0.3mm] (fcf.east) -- (lab.west);  
            
        \end{tikzpicture}
    }
    
    \caption{Two steps of the proposed approach to solve Image Clustering. Generating multiple ``artificial views" of the original data from several CNNs improves the clustering results by complementarity of the feature representations. Solving Multi-View Clustering end-to-end (DMVC) improves MVC results while generating a new representation that is low-dimensional and compact.}
    \label{fig:method}
\end{figure}

\section{Deep Multi-View Clustering}

\subsection{Preliminaries: Deep end-to-end clustering}\label{sec:prelim_e2e}

End-to-end clustering methods based on neural networks have produced excellent results in the past two years. A complete literature review of the topic is outside the scope of this paper, for that we refer the reader to the following recent survey \cite{survey_e2e}. However, we describe  two classes of methods relevant to our approach in more detail. 

In \cite{jule}, the authors introduce a connectivity-based clustering method called Joint Unsupervised Learning of deep representations and image clusters (JULE). This approach learns a new representation of the initial features with a neural network. The training of the neural network is embedded within the generic agglomerative clustering framework. The algorithm iteratively merges clusters and trains the neural network to minimize a clustering loss based on the affinity computed from the K-nearest neighbors of each point. This approach is currently producing state-of-the-art results on many image datasets.

In \cite{dec}, the authors propose an end-to-end centroid-based clustering method, unsupervised Deep Embedding for Clustering analysis (DEC). This approach begins by pretraining a multi-layer perceptron (MLP) using an auto-encoder input reconstruction loss function. Then, the MLP is fine-tuned to output a set of cluster centers which define cluster assignments. Both the network optimization and the centroid optimization are based on the minimization of the Kullback-Leibler (KL) divergence between the current distribution of the features and an auxiliary target distribution, derived from high-confidence predictions. Improved Deep Embedding Clustering (IDEC), introduced in \cite{idec}, modifies DEC by replacing the loss function with a combined loss, which takes into account the auto-encoder reconstruction loss during fine tuning. This approach preserves the local structure of the data and appears to improve the DEC clustering results.

\subsection{Deep multi-view clustering (DMVC)}

In this section, we define our approach for solving MVC. Let $\mathcal{C}_{\text{ee}}$ be any deep clustering framework. $\mathcal{C}_{\text{ee}}$ is defined by a loss function $\mathcal{L}$ and a procedure $\mathcal{P}$ to optimize the loss function. Multiple approaches have already been adopted to define the clustering-oriented loss $\mathcal{L}$ and the optimization procedure $\mathcal{P}$ (examples are discussed in Section~\ref{sec:prelim_e2e}). Applying $\mathcal{C}_{\text{ee}}$ to a new unsupervised dataset $\mathcal{D}$ requires one to specify a function $f_{\theta}$, parameterized by $\theta$, which transforms $\mathcal{D}$ into a new feature space $\mathcal{D}_{\theta} = f_{\theta}(\mathcal{D})$. From there, $\mathcal{C}_{\text{ee}}$ applies $\mathcal{P}$ to minimize $\mathcal{L}(\theta, \mathcal{D})$, producing both a good representation $\mathcal{D}_{\theta_{\text{final}}}$ and a set of cluster assignments $y_{\text{final}}$. 

\vspace{-10pt}
\paragraph{\textcolor{bmv@sectioncolor}{Architecture}}
In general, $f_{\theta}$ is selected to be a neural network and the choice of the architecture depends on the kind of dataset. For example, when dealing with large images, $f_{\theta}$ can be a CNN and when $\mathcal{D}$ is composed of smaller vectors, $f_{\theta}$ can be a multi-layer perceptron (MLP). In the case of MVC, each element of $\mathcal{D}$ is a collection of vectors. For example, the $k^{th}$ element of $\mathcal{D}$ is written as $\{V_{i, k}, ~ \forall i \in [1, ... m]\}$. For this reason, to embed MVC into a deep clustering framework, we need to define a different neural network architecture for $f_{\theta}$, which we call MVnet. MVnet consists of a set of $m$ independent MLPs, denoted $mlp = \{mlp_1, ... mlp_m\}$, such that, $\forall i \in [1, ... m]$, the dimension of the input layer of $mlp_i$ is equal to the dimension of the output layer of the associated $FE_i$. We also define $mlp_{\text{out}}$, another MLP with input layer dimension equal to the sum of the dimensions of the output layers over the elements of $mlp$. Thus, an MVnet is composed of three layers: a parallel layer containing all the elements of $mlp$, followed by a concatenating layer which feeds into $mlp_{\text{out}}$. A visual representation of the MVnet architecture can be seen in the red box in Figure~\ref{fig:method}. We note that all the elements of $mlp$ are independent and do not share any weights.

\vspace{-10pt}
\paragraph{\textcolor{bmv@sectioncolor}{Training}}
DMVC is a generic framework and MVnet can be optimized using most deep end-to-end clustering approaches. Given a deep clustering framework, MVnet should be trained on the dataset $\mathcal{V}$, from which sample $K$ is $\mathcal{V}[K] = \{V_{i, K}, i \in [1, ... m]\}$. In our experience, training MVnet from scratch does not provide good results. Instead, it is better to pretrain each $mlp_i$ on $V_i$ using a deep clustering framework. Then, pretrain $mlp_{out}$ on the concatenation of the new feature representations extracted by the pretrained members of $mlp$. Finally, after initializing its weights with the appropriate pretrained MLPs, MVnet can be refined end-to-end on $\mathcal{V}$ to improve both the feature representation and the clustering results.

\vspace{-10pt}
\paragraph{\textcolor{bmv@sectioncolor}{Clustering}}
The method for obtaining the final cluster assignments depends on the deep clustering framework selected. For example, using DEC, the optimization procedure will output a set of centroids in the new feature space, which can straightforwardly be used to assign a cluster to each image. Using JULE, cluster assignments are generated during training. Data points are gradually grouped together to form clusters while $f_{\theta}$ is being trained.
\section{Experimental setup}\label{sec:exp}

\subsection{Datasets}
We use three different image datasets to validate our work. \emph{COIL100}~\cite{coil100} is composed of multiple images of the same objects from different angles. Images are centered around the object and background is neutral. \emph{UMist}~\cite{umist} is a facial recognition database. For each person, multiple pictures are taken under different light conditions and orientations. Each image is centered around the face with a neutral background. \emph{VOC2007}~\cite{voc2007} is an image classification dataset presenting visual objects from various classes in a realistic scene. This is a very challenging dataset for clustering: objects are not pre-segmented, backgrounds are complex, and  the images are quite large. Data statistics can be found in Table~\ref{table_data}.

\begin{table}[!ht]
\caption{Statistics of the datasets used for our experiments.} 
\label{table_data}
\begin{threeparttable}
\renewcommand{\TPTminimum}{\linewidth}
\vspace{\baselineskip}
\makebox[\linewidth]{
\begin{tabular}{cccc}

Dataset & COIL100 & UMist & VOC2007\tnote{1}
\tabularnewline \hline

\# Images & 7200 & 575 & 2841 \tabularnewline
\# Classes & 100 & 20 & 20 \tabularnewline
Image Size\tnote{2}
& 128x128 & 112x92 & Variable \tabularnewline
\end{tabular}}

\begin{tablenotes}
\item[1] \small \textit{We use a modified version of the VOC2007 test set. All the images presenting two or more labels have been removed in order to have ground truth to evaluate clustering quality.}
\item[2] \small \textit{The images must be preprocessed to match the input sizes of the CNN used ($224 \times 224$  for ResNet50 and VGG; $299 \times 299$ for Inception and Xception). This can be done with anti-aliasing interpolation.}
\end{tablenotes}
\end{threeparttable}
\end{table}

\subsection{Other methods for comparison}

\subsubsection{DMVC evaluation}
We implemented several different versions of the proposed DMVC approach. First, we consider a framework in which the MVnet weights are fixed after initialization (no end-to-end fine-tuning). We denote this DMVC-fix. DMVC-fix is implemented with two different end-to-end clustering frameworks, JULE~\cite{jule} and IDEC~\cite{idec}. Finally, the full DMVC pipeline is implemented within the JULE framework, which seems to perform the best on every dataset. This enables usto evaluate the influence of end-to-end training on multi-view datasets.

To evaluate the DMVC framework, we compare it against two other MVC methods. Let $\mathcal{C}$ stand for any clustering algorithm. The first naive approach consists in concatenating the different views and applying $\mathcal{C}$ on the merged representation. We denote this approach CC (concatenate + cluster). The second approach is based on ensemble clustering and is derived from~\cite{multiview_ensemble}. Here we generate a set of partitions $\mathcal{P} = \{P_1, ..., P_m\}$ by applying $\mathcal{C}$ on each element of $\mathcal{V}$. Then, the co-association matrix (CAM) of $\mathcal{P}$ is built, measuring how many times any pair of elements have been clustered together. The CAM is then used as a graph-distance matrix by any connectivity-based clustering algorithm (spectral clustering, agglomerative clustering, etc.). In our result tables, this approach is referred to as Multi-View Ensemble Clustering (MVEC). MVEC has been shown to produced state-of-the art results on MV clustering recently~\cite{multiview_ensemble}.

\subsubsection{Multiple CNN feature extraction evaluation}
For any pretrained CNN, we also report the result of applying $\mathcal{C}$ to the single set of features it extracts. This is informative in evaluating the impact of multi-view generation. For each approach, both JULE and IDEC are used for $\mathcal{C}$. KMeans (KM)~\cite{kmeans++} and Agglomerative Clustering (AC)~\cite{agglomerative}, two standard clustering methods, are also applied to every fixed CNN, CC and MVEC. Using simple clustering methods allows us to evaluate how multiple CNNs can benefit image clustering independently of other factors.
    
\subsection{Practical implementations}

In our experiments, we used the Keras implementations and pretrained weights~\cite{keras} of five CNN architectures: two VGG architectures (VGG16 and VGG19~\cite{vgg}), one ResNet architecture (ResNet50~\cite{resnet}), and two Inception-like architectures: (InceptionV3~\cite{inception} and Xception~\cite{xception}). When we discuss features extracted from a given CNN, we always refer to the activation just before the final linear layer for ImageNet classification. This layer is an average pooling layer for InceptionV3, Xception and ResNet50, and a fully-connected layer for both VGG architectures.
    
DMVC is a framework for \emph{unsupervised} classification, hence, we should not do any hyperparameter tuning. In all of our experiments, we use default parameters for every sub-algorithm used. For both KM and AC we use the default configuration of the scikit-learn implementations~\cite{sklearn}. For JULE and IDEC, we use the hyperparameters recommended by the authors (see original papers for more details). We only increase the learning rate to $5\times10^2$ for the MVnet fine tuning, as this appears to be consistently better for all the three datasets. Finally, for ensemble clustering, the co-association matrix is clustered with agglomerative clustering with average linkage. For IDEC, we generate centroids with a multilayer perceptron (MLP) with dimensions $d-500-500-2000-N$, where $d$ is the extracted feature vector dimension and $N$ is the number of classes that we are searching for. For JULE, the representation learning network is a MLP with dimensions $d-160-160$. For both methods, hidden layer activations are rectified linear units. The building blocks of MVnet are of the shapes defined above.

The clustering results are evaluated using normalized mutual information (NMI), which is commonly used in unsupervised classification. NMI ranges between $0$ and $1$, with $1$ representing perfect accuracy.

 \section{Experimental results}
 
 \subsection{Clustering results}
    
The clustering results on the three datasets are reported in Table~\ref{table_result}. Table~\ref{table_bf} reports the average results over the three datasets for each method. Computing averages makes sense in an unsupervised setting because, as cross-validation is impossible, a clustering pipeline needs to perform well on every dataset. BFN is the best fixed network over all datasets.

\begin{table}[!ht]
\caption{Clustering performances (NMI) for different IC methods. The proposed multi-view generation methods is evaluated by comparing the five first rows (independent CNN feature extractors) and the four last ones (multi-CNN features). The proposed end-to-end method to solve MVC is evaluated by comparing DMVC to CC and MVEC.\vspace{4mm}} 
\label{table_result}
\centering

\scalebox{0.75}{
\begin{tabular}{c|cccc|cccc|cccc}
    & \multicolumn{4}{c|}{\textbf{VOC2007}} & \multicolumn{4}{c|}{\textbf{COIL100}} & \multicolumn{4}{c}{\textbf{UMist}}\tabularnewline

    & JULE & IDEC & KM & AC & JULE & IDEC & KM & AC & JULE & IDEC & KM & AC\tabularnewline \hline
    VGG16 & 0.687 & 0.666 & 0.660 & 0.665 & 0.989 & 0.963 & 0.939 & 0.956 & 0.920 & 0.771 & 0.707 & \textcolor{Blue}{0.755}\tabularnewline
    VGG19 & 0.684 & 0.677 & 0.663 & 0.649 & 0.994 & 0.963 & 0.941 & 0.948 & 0.933 & 0.742& \textcolor{Blue}{0.729} & 0.729 \tabularnewline
    InceptionV3 & 0.768 & 0.760 & 0.620 & 0.675 & 0.984 & 0.957 & 0.942 & 0.953 & 0.823 & 0.705 & 0.646 & 0.692 \tabularnewline
    Xception & 0.759 & \textcolor{Blue}{0.779} & 0.668 & \textcolor{Blue}{0.720} & 0.986 & 0.955 & 0.933 & 0.955 & 0.829 & 0.707 & 0.591 & 0.678 \tabularnewline
    ResNet50 & 0.679 & 0.691 & 0.681 & 0.649 & \textcolor{Blue}{\textbf{0.997}} & \textcolor{Blue}{0.973} & \textcolor{Blue}{0.962} & \textcolor{Blue}{0.967} & 0.919 & \textcolor{Blue}{0.784} & 0.686 & 0.723 \tabularnewline \hline
    CC & 0.718 & 0.587 & \textcolor{Blue}{0.698} & 0.698 & 0.995 & 0.886 & 0.944 & 0.952 & 0.855 & 0.699 & 0.681 & 0.700 \tabularnewline
    MVEC & 0.785 & \textcolor{Blue}{0.782} & \textcolor{Blue}{0.728} & \textcolor{Blue}{0.741} & \textcolor{Blue}{\textbf{0.996}} & \textcolor{Blue}{0.977} & \textcolor{Blue}{0.958} & \textcolor{Blue}{0.967} & \textcolor{Blue}{\textbf{0.963}} & \textcolor{Blue}{0.797} & \textcolor{Blue}{0.748} & \textcolor{Blue}{0.761} \tabularnewline
    DMVC-fix & \textcolor{Blue}{\textbf{0.792}} & 0.730 & N/A & N/A & \textcolor{Blue}{\textbf{0.996}} & \textcolor{Blue}{0.973} & N/A & N/A & \textcolor{Blue}{\textbf{0.963}} & 0.737 & N/A & N/A \tabularnewline
    DMVC & \textcolor{Blue}{\textbf{0.810}} & - & N/A & N/A & 0.995 & - & N/A & N/A & \textcolor{Blue}{\textbf{0.971}} & - & N/A & N/A  
\end{tabular}}

\begin{tablenotes}
\item \small \textit{The two best methods for each column are in blue, the two best overall methods are in bold. N/A is for incompatible methods.}
\end{tablenotes}
\end{table}

\noindent The method proposed in this paper is composed of two steps: multi-CNN feature extraction and deep multi-view clustering. Each of these contributions can be leveraged separately or jointly, thus the results are discussed as follows: 
\begin{itemize}
\setlength\itemsep{0.01em}
    \item In Section \ref{sec:eval_mv}, we discuss the benefits of using several feature extractors instead of one for IC,
    \item In section \ref{sec:eval_dmvc} and \ref{sec:representation}, we discuss the advantages of using an end-to-end approach to address MVC,
    \item In Section \ref{sec:eval_combined} and \ref{sec:representation}, we explain the upsides of combining the two approaches to solve IC.
\end{itemize}

\subsubsection{IC can be improved by using features extracted from several pretrained CNNs}\label{sec:eval_mv}

Methods representing the data with different views extracted from multiple CNNs consistently outperform every method with a fixed feature extractor. This is also true with KM and AC, which supports the proposed way to solve image clustering as multi-view clustering, even before refinement. The only exception is for the COIL100/ResNet50 combination, but the results are similar. Moreover, NMI is very close to 1 for this configuration, which possibly indicates that the difference in performance derives from a few outliers. The importance of multi-view generation can also be observed in Table~\ref{table_bf}. The scores reported in Table~\ref{table_bf} represent how well each method performs ``in general,'' and show that there is no fixed feature extractor pipeline which is consistently better than a MV method across multiple datasets. 

\begin{table}[!ht]
\caption{Average NMI scores over the three datasets using JULE. The Best Fixed Network (BFN) represents the single network which performs best on average. As the feature extractor cannot be specifically selected for a given dataset, comparing multi-view results to BFN is relevant.}
\label{table_bf}
\centering
\vspace{\baselineskip}
    \begin{tabular}{c|cccccc}
        
        Clustering routine & BFN & CC & MVEC & DMVC-fix & DMVC \tabularnewline \hline
        Average NMI score & 0.870 & 0.856 & 0.915 & 0.917 & \textbf{0.925}
        
    \end{tabular}
  
\end{table}

Such results validate the hypothesis that different pretrained architectures contain complementary information about a new unsupervised dataset and \emph{justify the use of multi-view generation from different CNN architectures} rather than a fixed CNN for feature extraction. In particular, for VOC2007, gathering results from multiple CNNs using MVEC or DMVC is highly beneficial, which suggests that the proposed MV framework can improve clustering for complex realistic datasets. Finally, it is interesting to note that, most of the time, CC performs worse than fixed networks. This means that features extracted from each CNN should first be processed independently before being used for clustering. This gives an experimental justification for why subnetworks of MVnet should first be pretrained separately.

\subsubsection{MVC can be improved by adopting an end-to-end approach}\label{sec:eval_dmvc}

For both VOC2007 and UMist, fine tuning MVnet using the JULE framework improves the clustering results over the DMVC with no end-to-end retraining (DMVC-fix), which \emph{validates the use of an end-to-end approach for MVC}. Also, apart from COIL100, where results are very similar, DMVC outperforms MVEC. For COIL100, DMVC does not improve clustering over ResNet50. A possible explanation is that ResNet50 already separates the clusters well (NMI = 0.997) and there is not much left to be learned. Another reason for using deep methods for solving MVC is discussed in Section~\ref{sec:representation}, where it is shown that solving MVC end-to-end generates a more compact, unified representation.

\subsubsection{Combining multiple CNNs multi-view generation and DMVC produces state-of-the-art results at IC}\label{sec:eval_combined}

To the best of our knowledge, the results reported in Table~\ref{table_result} represent \emph{the new state-of-the-art for clustering these datasets}. The use of multiple pretrained feature extractors, combined with the proposed DMVC framework, enables us to outperform other approaches on the tested IC problems. The results indicate that, when facing a new unsupervised image classification dataset without any specific information about the data, we recommend that one adopt the approach proposed in this paper: transform the problem into MVC from all available pretrained CNNs and use DMVC (trained within the JULE framework) to solve it.

\subsection{Learned representations}\label{sec:representation}

The quality of a clustering algorithm can also be evaluated by the quality of the new feature representation it generates. To analyze the quality of the feature representation extracted with DMVC, we re-cluster several feature representations, from different stages of the DMVC pipeline, using KM. KM is a simple clustering algorithm which performs better on representations presenting compact clusters, which are distant from each others. We choose InceptionV3 to represent the fixed CNN feature representation methods as it performs best on VOC2007. Results are reported in Table~\ref{table_representation} and clearly demonstrate that we generate better features as we progress through the DMVC pipeline.

\begin{table}[!ht]
\caption{Comparison of clustering performance (NMI) of KMeans applied to different representations of the dataset. If a feature representation gets a high score, it means that it presents compact clusters which are distant from each others.} 
\label{table_representation}

\centering
\vspace{10pt}
    \begin{tabular}{c|ccc}

        & VOC2007 & COIL100 & UMist \tabularnewline \hline
        InceptionV3 & 0.624 & 0.932 & 0.680 \tabularnewline
        InceptionV3 + JULE & 0.754 & 0.938 & 0.775 \tabularnewline
        DMVC-fix & 0.759 & 0.961 & 0.895 \tabularnewline
        DMVC & \textbf{0.786} & \textbf{0.964} & \textbf{0.973} \tabularnewline
    
    \end{tabular}
    
\end{table}

\begin{figure}[!ht]
    \centering
    
    \begin{subfigure}[b]{0.23\textwidth}
        \centering \includegraphics[width=\textwidth]{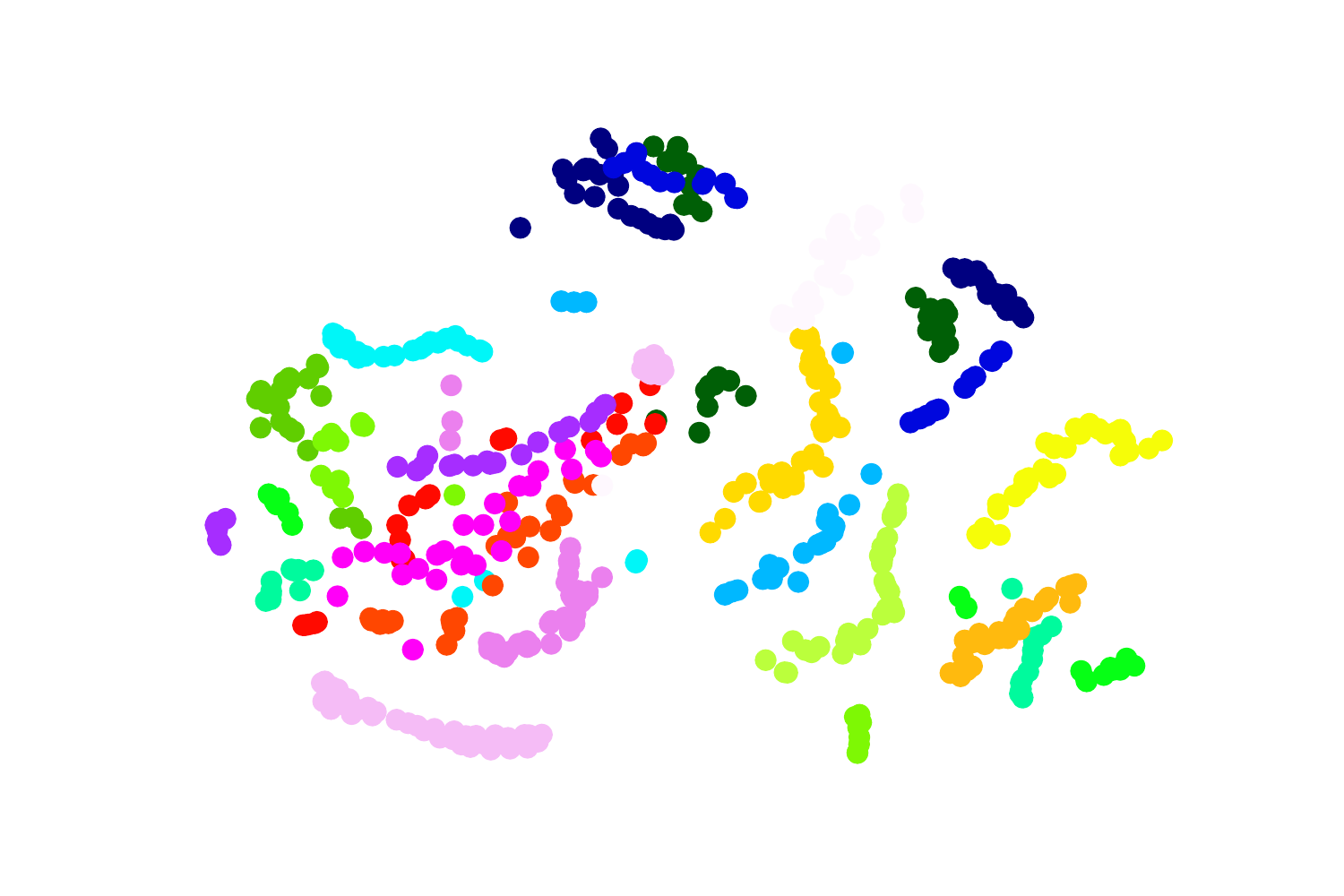}
        \caption{\footnotesize InceptionV3 features}
    \end{subfigure}
    ~
    \begin{subfigure}[b]{0.23\textwidth}
        \centering \includegraphics[width=\textwidth]{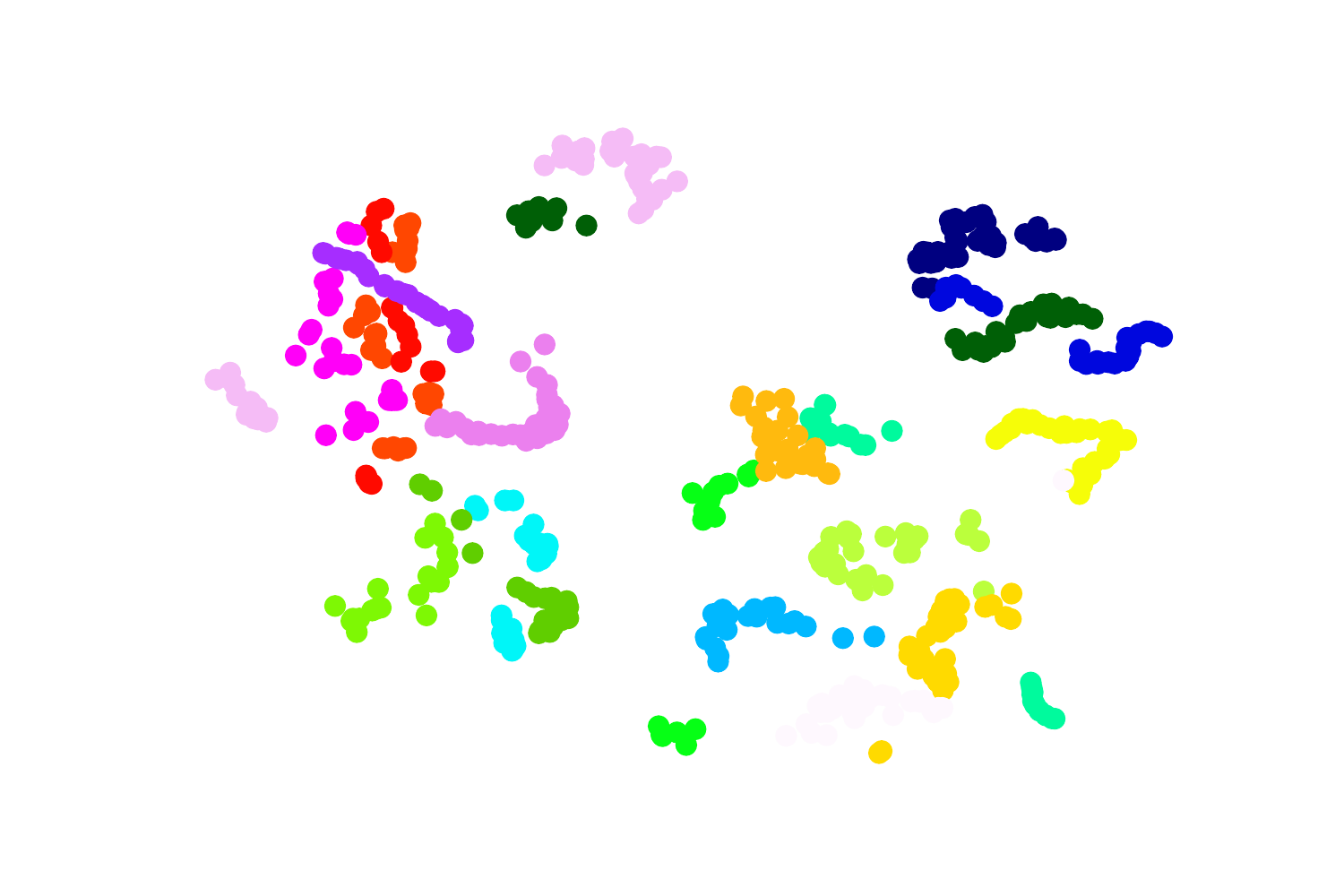}
        \caption{\footnotesize InceptionV3 + JULE}
    \end{subfigure}
    ~
    \begin{subfigure}[b]{0.23\textwidth}
        \centering \includegraphics[width=\textwidth]{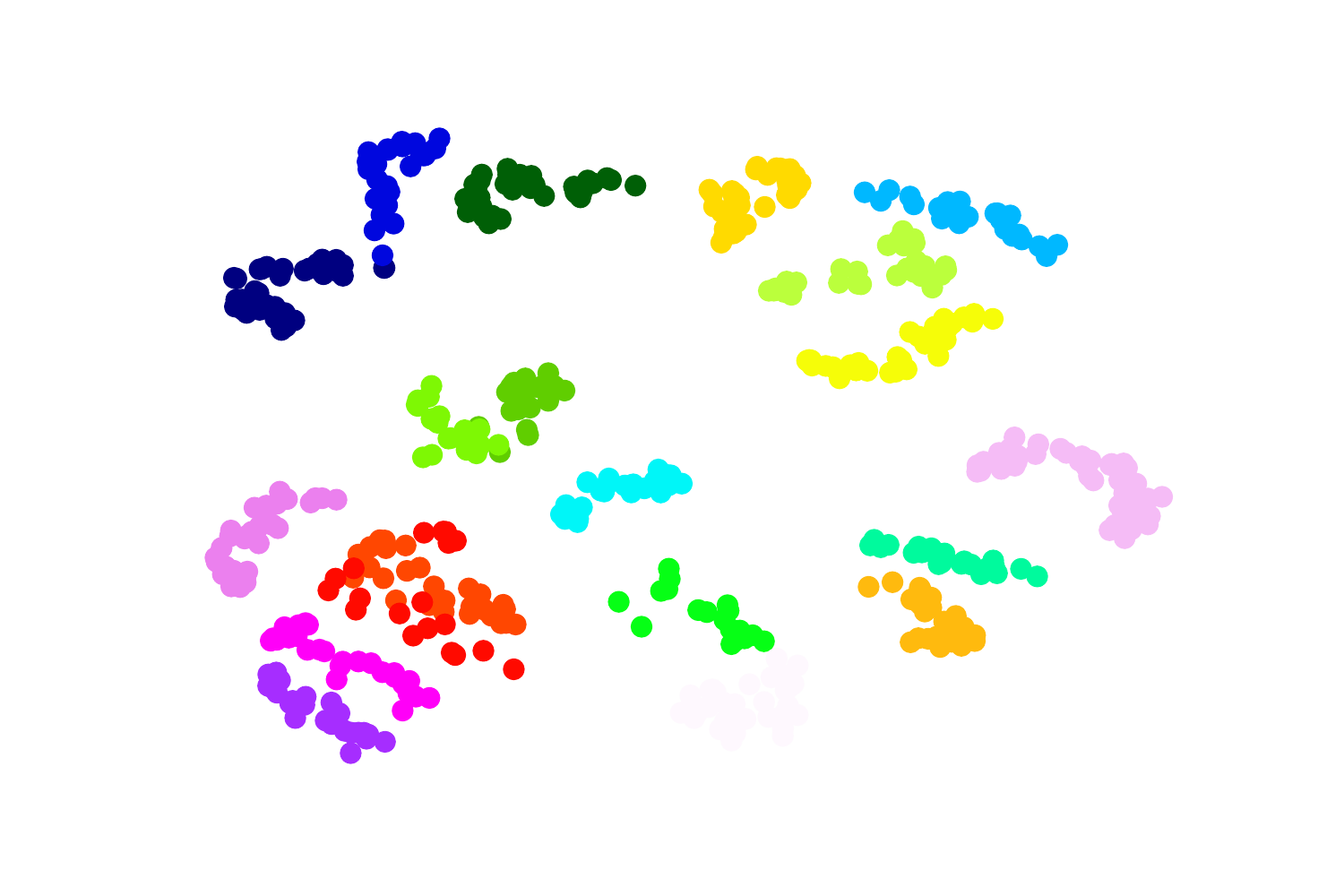}
        \caption{\footnotesize DMVC-fix}
    \end{subfigure}
    ~
    \begin{subfigure}[b]{0.23\textwidth}
        \centering \includegraphics[width=\textwidth]{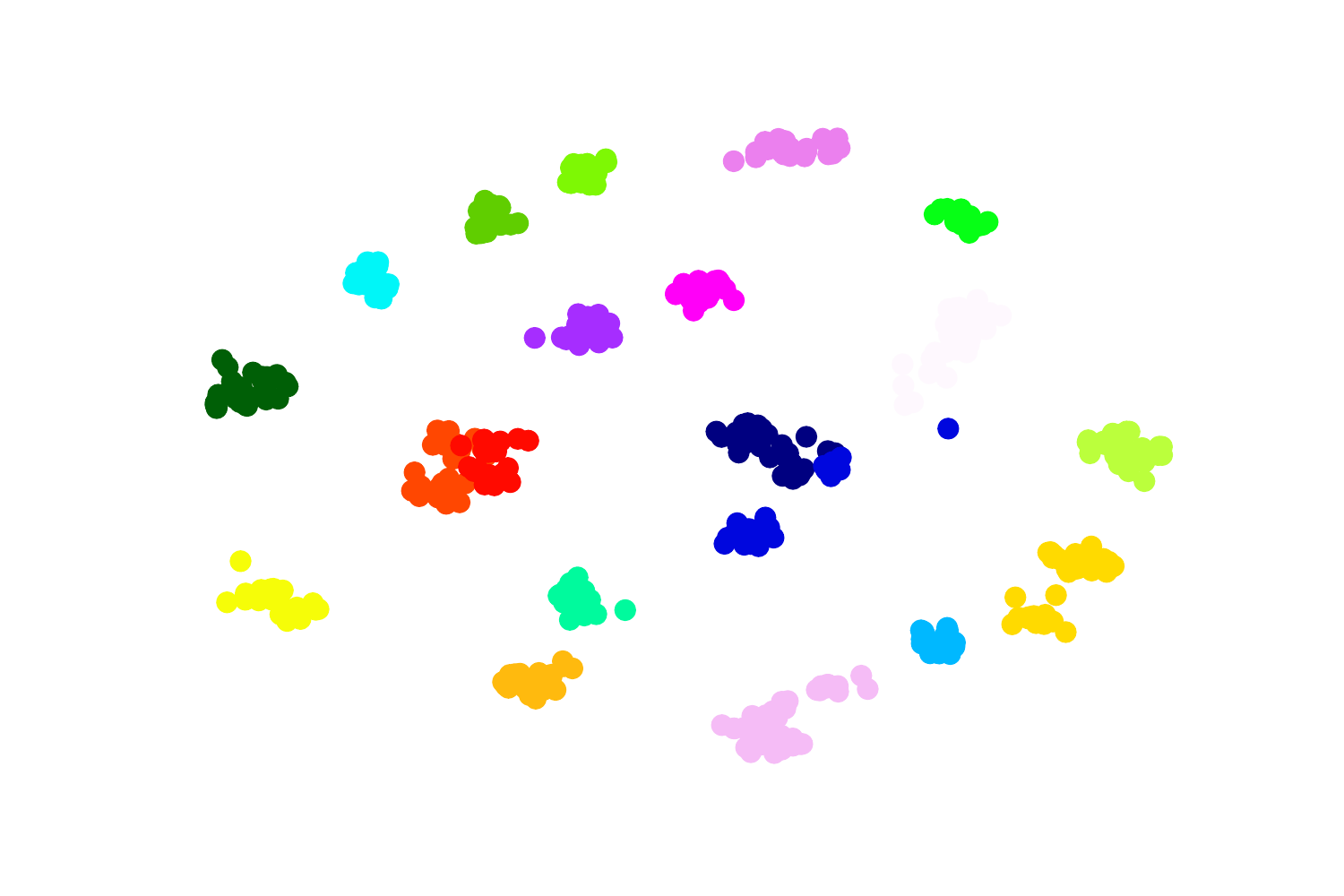}
        \caption{\footnotesize DMVC}
    \end{subfigure}
    
    \caption{\textit{Best viewed in color}. 2D t-SNE visualization of the features extracted from the UMist dataset at different stages of the DMVC framework. The feature representation becomes more compact and better separated as we progress in the DMVC pipeline.}
    \label{fig:visualization}
\end{figure}

\noindent Figure \ref{fig:visualization} provides a visual way to evaluate DMVC features. It is a 2d t-SNE representation of different features at different stages of DMVC for the UMist dataset. Comparing subfigures (b) and (c), we see that using multiple CNNs enables to obtain a better separated feature representation of the dataset. Comparing subfigures (c) and (d), we see that fine tuning DMVC end-to-end produces representations that generate more compact clusters.

\section{Conclusion}

We propose a two-step approach to solving the image clustering problem. First, we generate multiple representations of each image using pretrained CNN feature extractors, and reformulate the problem as a multi-view clustering problem. Second, we define a multi-input neural network architecture, MVnet, which is used to solve MVC in an end-to-end manner using any deep clustering framework. Implementing this pipeline with JULE is state-of-the-art and sets a new benchmark for image clustering on the datasets presented. This approach also has the advantage of removing the design choice of selecting a single feature extractor. 
    
Our experimental results illustrate that different CNNs, pretrained on the same task, may contain different and complementary information about a dataset.  Differences may arise from a number of sources including the architecture (number of layers, layer shape, presence of skip connections, etc.), regularization methods, or loss functions used for training. Investigating which parameters influence knowledge transfer to unsupervised tasks is an interesting axis of research for future work.   
Finally, we note that pretrained CNNs are used as feature extractors for many applications, not just clustering. 
Using multiple pretrained CNNs to define a multi-view learning problem may be appealing for other tasks where complementary information present in pretrained feature extractors may improve performance. 

\section*{Acknowledgments}
This work was carried out under a Fulbright Haut-de-France Scholarship as well as a scholarship from the Fondation Arts et M\'etiers (convention No. 8130-01). This work is also supported by the European Union's 2020 research and innovation program under grant agreement No.688807, project ColRobot (collaborative robotics for assembly and kitting in smart manufacturing).

\bibliography{bmvc.bbl}
\end{document}